\documentclass[10pt,twocolumn,letterpaper]{article}

\usepackage{wacv}
\usepackage{times}
\usepackage{epsfig}
\usepackage{graphicx}
\usepackage{amsmath}
\usepackage{amssymb}
\usepackage{booktabs}
\usepackage{bm}
\usepackage{multirow}

\def\wacvPaperID{205} 

\wacvalgorithmstrack   

\wacvfinalcopy 

\ifwacvfinal
\usepackage[breaklinks=true,bookmarks=false]{hyperref}
\else
\usepackage[pagebackref=true,breaklinks=true,colorlinks,bookmarks=false]{hyperref}
\fi

\newcommand{\lossL}{\mathcal{L}}
\usepackage[capitalize]{cleveref}
\crefname{section}{Sec.}{Secs.}
\Crefname{section}{Section}{Sections}
\Crefname{table}{Table}{Tables}
\crefname{table}{Tab.}{Tabs.}

\begin{document}

\title{Dynamic Template Initialization for Part-Aware Person Re-ID}

\author{Kalana Abeywardena \and
Shechem Sumanthiran \and
Sanoojan Baliah  \and 
Nadarasar Bahavan \and 
Nalith Udugampola  \and 
Ajith Pasqual \and 
Chamira Edussooriya \and 
Ranga Rodrigo \and 
\\
Department of Electronic and Telecommunication Engineering \\University of Moratuwa, Sri Lanka \\
\tt\small \{kalanag,shechems,sanoojanb,tbahavan,nalithl,pasqual,chamira,ranga\}@uom.lk
}


\maketitle
\thispagestyle{empty}

\begin{abstract}
    Many of the existing Person Re-identification (Re-ID) approaches depend on feature maps which are either partitioned to localize parts of a person or reduced to create a global representation.
    While part localization has shown significant success, it uses either na\"ive position-based partitions or static feature templates. These, however, hypothesize the pre-existence of the parts in a given image or their positions, ignoring the input image-specific information which limits their usability in challenging scenarios such as Re-ID with partial occlusions and partial probe images.
    In this paper, we introduce a spatial attention-based Dynamic Part Template Initialization module that dynamically generates part-templates using mid-level semantic features at the earlier layers of the backbone. 
    Following a self-attention layer, human part-level features of the backbone are used to extract the templates of diverse human body parts using a simplified cross-attention scheme which will then be used to identify and collate representations of various human parts from semantically rich features, increasing the discriminative ability of the entire model.
    We further explore adaptive weighting of part descriptors to quantify the absence or occlusion of local attributes, and suppress the contribution of the corresponding part descriptors to the matching criteria. 
    Extensive experiments on holistic, occluded, and partial Re-ID task benchmarks demonstrate that our proposed architecture is able to achieve competitive performance. Codes will be included in the supplementary material and will be made publicly available.
\end{abstract}

\section{Introduction}

Person Re-identification (Re-ID) has fast become one of the most critical vision tasks in the surveillance domain, especially due to its mission-critical application in video surveillance, autonomous driving, and activity analysis \cite{zhang2018learning, zheng2012reidentification, yang2014salient, liao2015person}. Re-ID is a task in which a probe image is used to search for a person among a gallery of person images cropped from a larger scene \cite{gong_xiang_2011, zheng2016person}.
Although Re-ID performance has recently improved, certain inherent challenges still require solutions, especially when dealing with persons being occluded in cluttered scenes. 

Humans utilize different visual cues to Re-ID in different contexts: for quick re-identification, humans utilize vivid global features that include the color of clothing and large accessories associated with the person; whereas in more challenging environments, people pay attention to discriminating and unique local features such as the features of a face \cite{wolfe1992parallel}. 
This illustrates that awareness of local features unique to a person can be beneficial in successful Re-ID. However, designing algorithms that are aware of local features can be challenging as they need to learn subtle changes to discriminate successfully among different persons.  

In recent works, many solutions have been proposed to learn discriminative local features. These can be generally divided into three categories: (1) hand-crafted splitting-based methods that divide and extract part features from the local patches \cite{he2018deep, zheng2015partial_Partial-REID, he2018recognizing} and stripes \cite{fan2018scpnet, sun2018beyond, qi2021exam} of the image or the feature map; (2) extra semantic-based methods that directly utilize either human parsing \cite{he2019foreground, he2020guided} or pose estimation models \cite{miao2019pose_Occludded-Duke, wang2020high, gao2020pose}; and (3) attention-based methods \cite{sun2019perceive, zhuo2019novel} that use attention schemes to localize discriminative human parts. 
However, each of these methods has inherent flaws. Hand-crafted splitting methods introduce additional background noise and poor alignment of the human parts due to the coarse features extracted. Human pose/parsing methods heavily rely on the performance of the off-the-shelf human pose/parsing estimation models, which are trained on datasets of different domains. Further, the adoption of human part detectors in Re-ID is challenging due to the lack of extra annotations in Person Re-ID datasets. Attention schemes utilize static templates that are separately learnt during training to localize and identify which portions of the feature maps correspond to human body parts. Finally, the above methods strongly assume that every body part is present in every image, disregarding the possibility of occlusions or partial images.

To address these flaws, we introduce a mechanism that learns part-based information using only the identity labels. Unlike the recent work that adopts either na\"ive position-based partitioning \cite{he2020guided, wang2020high} or static feature templates \cite{fan2018scpnet, sun2018beyond, li2021diverse}, we propose a method which dynamically initialize the part-aware spatial templates using input image-specific features extracted from the backbone model. We model the correlation between the extracted mid-level local features from the earlier layers of a backbone network that represents human part-level features and a set of learnable part-weights in our proposed \emph{Dynamic Part Template Initialization (DPTI) module}.
The acquired feature maps from the backbone model are first passed through a spatial self-attention layer to suppress background noise. The resulting attentive features are used to dynamically initialize part-aware templates. 
The proposed dynamic template generation enables the model to utilize mid-level features of the backbone as auxiliary information and  fully utilize the power of attention mechanism to obtain human part-templates. Unlike the static templates in \cite{li2021diverse}, where the learnt kernels are utilized to extract human part templates, our proposed dynamic templates do not make assumptions about the presence, location, or size of any body parts which enhances its capability to accurately identify the useful attributes from a given input image. 

By initializing part-templates dynamically, we do not rely on the extra annotations of each human body part. Instead, the templates are used to localize and aggregate discriminative features which relate to human parts from the high-level features of the last layer of the backbone using our proposed \emph{Part Attention Module}. This is achieved via a cross-attention scheme, resulting in the final part-descriptors used for matching persons. We are able to fully leverage the discriminative power of the cross-attention by using input-aware queries, as used in the transformer architectures for image classification \cite{kolesnikov2021image}. Therefore, the most prominent local features for a given pedestrian image will be extracted and converted into an embedding without manually altering the number of templates to be extracted based on the task as with static part-templates. To ensure every part embedding does not collapse to the same vector, we propose a modified diversity loss function from \cite{li2021diverse}. This encourages each template to learn different discriminative features about a pedestrian. Dynamic part-template initialization is useful for dealing with a number of challenges that are present not only in the Re-ID domain but in other applications as well.

We further explore an \emph{Adaptive Part Weighting} mechanism that weighs the part-descriptors using the part-template attention weights. We hypothesize the lower mean attention activation throughout the intermediate feature map indicates whether the part being localized is either absent or occluded. Thus, we quantitatively analyzed how it affects the person Re-ID, specially with partial probes and occluded images, by re-scaling the part-descriptors based on the mean activation of the part-template initialization module. We use the cosine distance metric with normalization of the feature vectors to ensure that the adaptive weighting reduces the impact of occluded or missing parts on the final distance.

The major contributions of our method can be summarized as follows:
\begin{enumerate}
    \item We propose a dynamic part-based spatial template initialization mechanism from intermediate feature maps of the backbone network to encourage template initialization based on less discriminative features.
    \item We propose to use a cross-attention mechanism between the dynamic templates and high-level discriminative feature maps to collate discriminative information of parts.
    \item We explore an adaptive weighting mechanism 
    in order to deal with occlusions and omissions of parts in the person images. 
\end{enumerate}
\section{Related Work}

\noindent \textbf{Person Re-Identification:} Person Re-ID aims to match complete person images across disjoint camera views \cite{zheng2015scalable, chen2017beyond, li2018harmonious, Chen_2019_ICCV} assuming all the human parts are visible in a given person image. Due to large intra-class and small inter-class variations caused by different viewpoints, poses illuminations, and occlusions, Re-ID is a challenging task. Existing methods for Re-ID can be grouped into hand-crafted descriptors \cite{liao2015person, matsukawa2016hierarchical}, metric learning methods \cite{koestinger2012large, xiong2014person}, and deep representation learning \cite{sun2018beyond, liu2018pose, sarfraz2018pose, su2017pose, song2018mask, kalayeh2018human, li2018harmonious, li2018unified}. Many recent state-of-the-art works utilize part-based features for holistic person Re-ID task using different approaches: human parsing models and assembling final discriminative representations with part-level features \cite{kalayeh2018human}; uniform partitioning of feature maps to learn part-level features by multiple classifiers \cite{sun2018beyond, qi2021exam}; and attention-based methods to extract part-level features \cite{zhao2017deeply, li2021diverse}. Building on attention-based part-aware models, we propose a DPTI module that creates local feature-aware templates in a self-supervised manner without any additional annotations or inputs to localize human parts within a given input-image. Further, our part templates are adaptive, rather than having static partitions of the input image or the outputs of the feature extractor.

\begin{figure*}[ht!]
    \centering
    \includegraphics[width=0.8\textwidth]{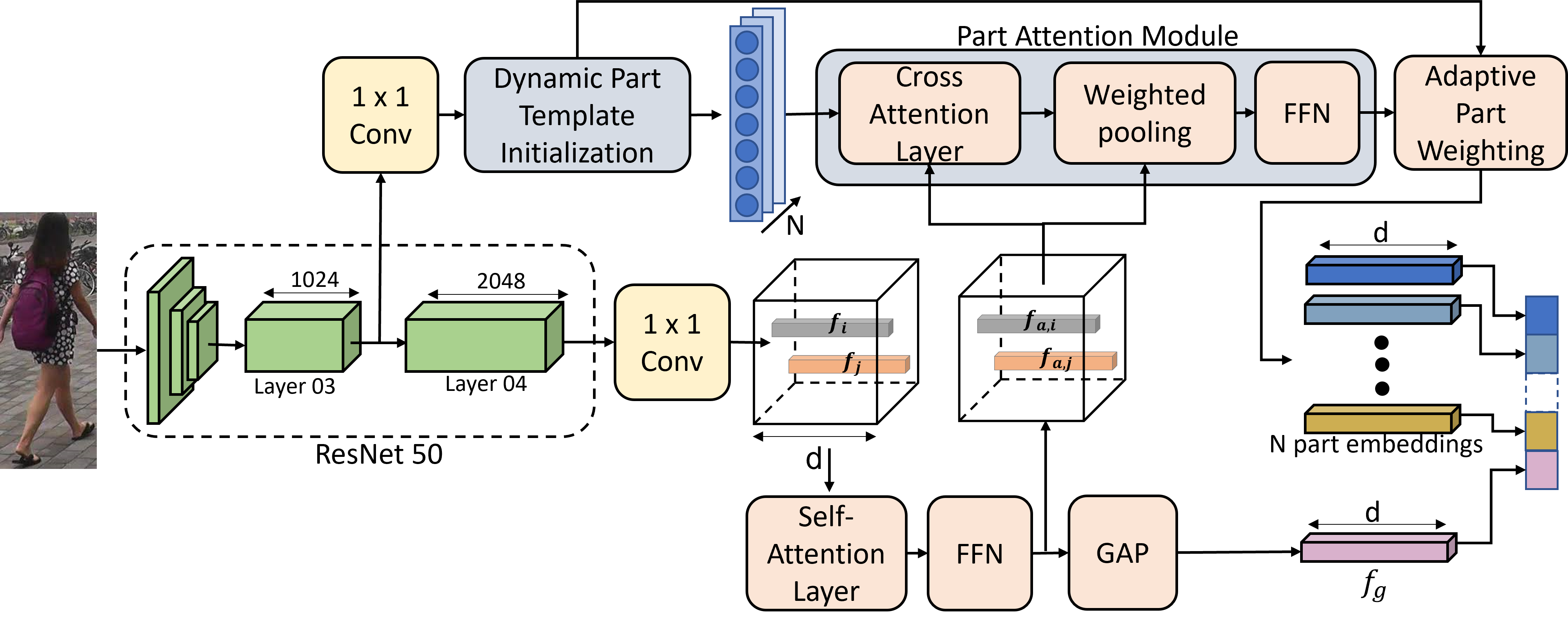}
    \caption{The proposed architecture: Intermediate feature maps are used by the DPTI module to compute part-templates. Feature maps from the final layer of the backbone are passed through a self-attention layer. The resulting attention outputs are collated by the Part Attention module to localize discriminative features of human parts. These are then weighted and combined with the global descriptor to form the final descriptor.}
    \label{fig:architecture}
\end{figure*}

\noindent \textbf{Occluded Person Re-ID:} 
Given occluded probe images, occluded person Re-ID aims to find the same person of full-body or occluded appearance in disjoint cameras. This task is more challenging due to incomplete information and spatial misalignment.
To deal with these challenges, some works have combined occluded/non-occluded binary classification loss \cite{zhuo2018occluded_Occluded-REID} to distinguish the occluded images from holistic ones.
Others reconstruct occluded feature maps and propose a spatial background-foreground classifier to reduce background clutter \cite{he2018deep}. Pose-guided feature alignment  \cite{miao2019pose} uses pose landmarks to mine discriminative parts and address occlusion noise. A pose-guided visible part matching model has been proposed to learn discriminative part features with pose-guided attention \cite{gao2020pose}.
In \cite{wang2020high}, the authors exploit graph convolutional layers to learn high order human part relations for robust alignment. Others transform the holistic image to align with the partial one and then calculate the distance of the aligned pairs \cite{luo2020stnreid}. Recently, \cite{li2021diverse} proposed part prototype-based transformer decoder to learn the part templates to identify the pedestrian parts.

Although the above methods can solve the occlusion problem to some extent, most of them heavily rely on off-the-shelf human parsing models or pose estimators. With our proposed method, we model the prominence of local features that are initialized as templates. 
Therefore, the occluded parts of a gallery image will be suppressed by the part-template initialization module, and have a low impact on Re-ID.

\noindent \textbf{Partial Person Re-ID:} Partial person images can occur due to imperfect detection of persons. Partial person Re-ID aims to match partial probe images to holistic gallery images. Attempts at addressing partial Re-ID include a global-to-local matching model to provide complementary spatial layout information \cite{zheng2015partial_Partial-REID}. An alignment-free approach that exploits the reconstruction error to reconstruct the feature map of a partial query from the holistic pedestrian was proposed in \cite {he2018deep}, and further improved by a foreground-background mask to avoid the influence of backgrounds clutter in \cite{he2019foreground}. STNReID that combines a spatial transformer network and a Re-ID network for partial Re-ID, was introduced in \cite{luo2020stnreid}. In \cite{sun2019perceive}, a Visibility-aware Part Model was introduced to perceive the visibility of part regions through self-supervision. In \cite{he2021partial}, a self-supervised learning framework that learns correspondence between image patches without any additional part-level supervision was proposed to recognize the partial input with the assistance of a part-part correspondence learning. A part-aware transformer network with a prototype transformer decoder that learns diversified part-based representation was proposed by \cite{li2021diverse}.

Our proposed approach is able to deal with partial images by dynamically initialize templates and weigh the corresponding parts based on the partial input image. Therefore, the architecture will only attempt to match, what is present and ignore the sections of the image that are absent.

\section{Proposed Architecture}

\subsection{Overview}

Our architecture is shown in Fig. \ref{fig:architecture}. There are 4 components of significance: feature extractor, DPTI module, part attention module and adaptive weighting module. 

The feature extractor used is a generic convolutional neural network (CNN). A person image is input into the feature extractor, which reduces the spatial dimensions and increases the channel dimension as the image propagates through the network. The DPTI module uses the feature maps extracted from part-way through the feature extractor, such as from the last layer of the penultimate stage, which represent the mid-level semantics of local attributes of person from the input image. The part templates are initialized from these feature maps using a cross-attention \cite{vaswani2017attention} mechanism. The part attention module takes the part templates as inputs, along with the feature maps from the final layer of the feature extractor. The feature maps are fed through a self-attention \cite{vaswani2017attention} encoder, and then a cross-attention network with the part templates is used as the queries. The outputs form the part descriptors of the person. 
The global descriptor and part descriptors are concatenated to form the final person descriptor. During training, this concatenated descriptor is jointly optimized using cross-entropy loss and triplet loss, along with a modified diversity loss to encourage differential part learning. The identity predictions are made using the descriptor, using a cosine distance metric to determine similarity between descriptors.

\subsection{Feature Extractor}

While any generic feature extractor can be used for the feature extractor, we utilize the popular ResNet-50 \cite{he2016deep}. To obtain larger feature maps at the final layer and to be consistent with previous works, we change the stride of the last layer from 2 to 1. 
At the input, each person image is resized to a fixed size of $H \times W$. The ResNet-50 architecture can be divided into 4 stages, each stage working on progressively smaller feature maps. The output feature maps of the final layer of the $l$-th stage is $\mathbf{F}_{l} \in \mathbb{R}^{n_{l} \times h_{l} \times w_{l}}$, where $n_{l}$ denotes the number of channels, $h_l$ is the height and $w_l$ is the width. We choose the stage $l < 4$ to be one of the first 3 stages. Similarly, the output feature maps of the final layer of the final stage is $\mathbf{F} \in \mathbb{R}^{n \times h \times w}$. 

\subsection{Dynamic Part Template Initialization Module}

Features extracted by the earlier layers of deep CNNs tend to be more generic and have lower-level semantics. These are generally not very discriminative and are more useful for later layers to localize more complex, discriminative features. Therefore, instead of using $\mathbf{F}$ to initialize the part templates, we choose feature maps $\mathbf{F}_{l}$. 

To create the templates, we first pass $\mathbf{F}_{l}$ through a $1 \times 1$ convolutional layer to reduce the feature dimensions to $d$. We obtain $\tilde{\mathbf{F}}_{l} \in \mathbb{R}^{d \times h_{l} \times w_{l}}$, where $d < n_{l}$, and pass it through a spatial self-attention encoder to suppress background noise in the feature maps. We reshape $\tilde{\mathbf{F}}_{l}$ to $d \times h_{l}w_{l}$ to treat each spatial location as a separate feature vector, such that $\tilde{\mathbf{F}}_{l} = [\bm{f}_{l, 1}, \bm{f}_{l, 2}, \dots, \bm{f}_{l, h_{l}w_{l}}]$, where $\bm{f}_{l, i} \in \mathbb{R}^{d \times 1}$ for $i = 1, \dots, h_l w_l$ is the feature vector of the $i$-th location. Then, the features are linearly projected to obtain the keys, queries and values as given by
\begin{align}
    \bm{q}_{l, i} &= \mathbf{W}_{l, q} \bm{f}_{l, i}, &
    \mathbf{K}_{l} &= \mathbf{W}_{l, k} \tilde{\mathbf{F}}_l, &
    \mathbf{V}_{l} &= \mathbf{W}_{l, v} \tilde{\mathbf{F}}_l, 
    \label{projection}
\end{align}
where $\mathbf{W}_{l, k}, \mathbf{W}_{l, q}, \mathbf{W}_{l, v} \in \mathbb{R}^{d \times d}$ matrices are the linear projections for the keys, queries, and values respectively. The attention outputs are subsequently calculated as
\begin{align}
    \bm{\beta}_{l, i} &= \mathrm{softmax} \left(
    \frac{\mathbf{K}_l^T \bm{q}_{l,i} }{\sqrt{d_k}} 
    \right),
    \label{attention 1}
    \\
    \bm{g}_{l, i} &= \mathbf{V}_l~\bm{\beta}_{l, i},
    \label{attention 2}
\end{align}
where $d_k$ is the scaling factor. The final attention outputs are given as
\begin{equation}
    \bm{a}_{l, i} = \mathbf{FFN}(\bm{g}_{l, i} + \bm{f}_{l, i}) + \bm{g}_{l, i},
    \label{ffn}
\end{equation}
where $\mathbf{FFN}(\cdot)$ is a feed-forward network of equal input and output dimensions. The matrix of attention outputs from the $l$-th intermediate layer is denoted as $\mathbf{A}_{l} = [\bm{a}_{l, 1}, \dots, \bm{a}_{l, h_l w_l}]$.

We initialize $N$ part templates for each input. The part templates are initialized by a simplified cross-attention mechanism, in which the learnt keys are represented by the linear projection matrix. Therefore, the part-templates, denoted as $\mathbf{T}_t = [\bm{t}_{1}, \dots, \bm{t}_{N}] \in \mathbb{R}^{d \times N}$, are obtained by 
\begin{align}
    \bm{\gamma}_{i} &= \mathbf{A}_{l}^T \bm{q}_{t, i},
    \label{part weights}
    \\
    \bm{t}_{i} &= \mathbf{W}_{t, v} \mathbf{A}_l \mathrm{softmax} \left(\bm{\gamma}_{i} \right),
\end{align}
where $\bm{q}_{t, i} \in \mathbb{R}^{d \times 1}, i =  1, \dots, N$ represent the learnt part-weights that serve as the queries. $\mathbf{W}_{t, v} \in \mathbb{R}^{d \times d}$ is the linear projection matrix for the template values. 

\begin{figure}
    \centering
    \includegraphics[width=0.8\linewidth]{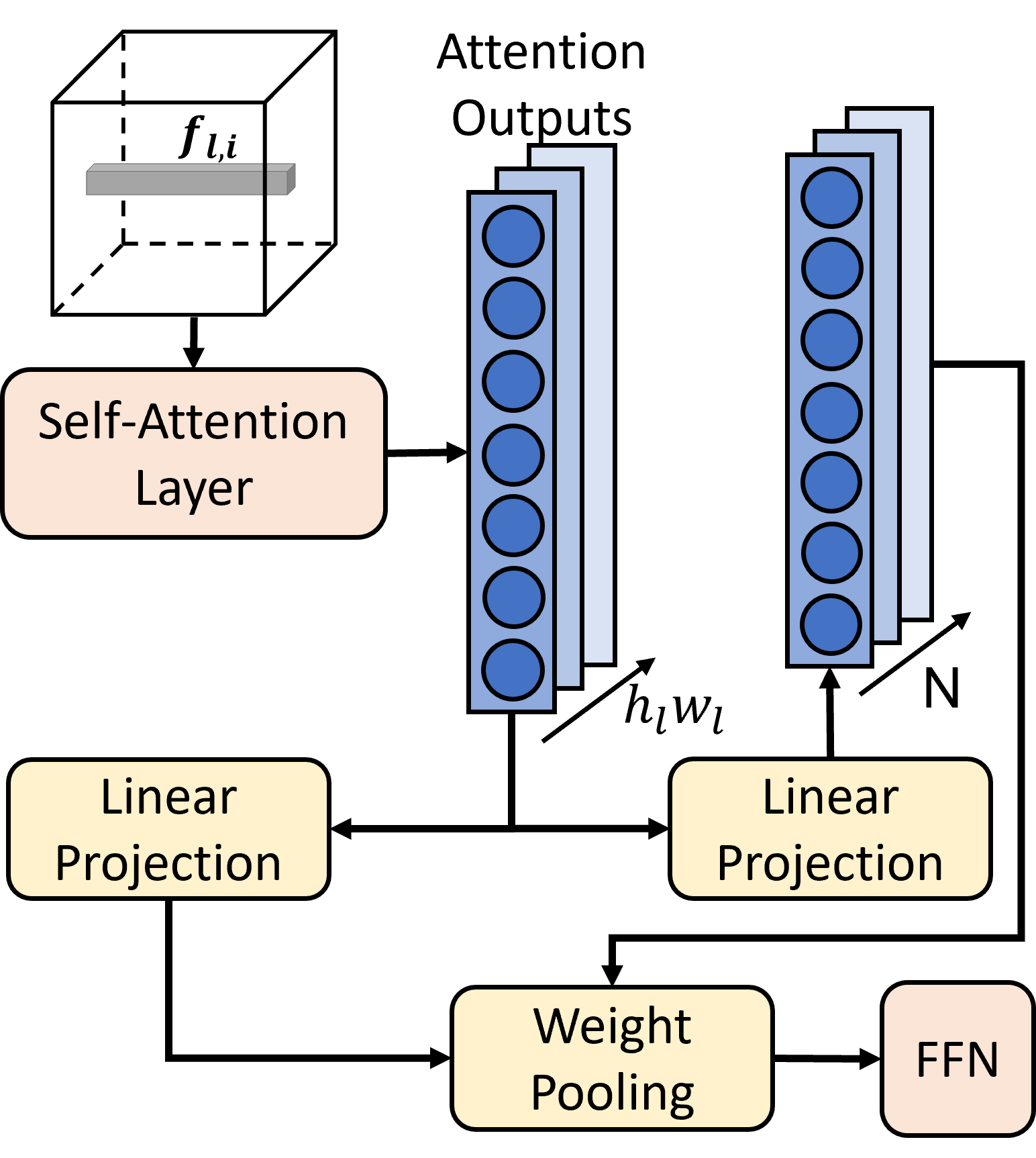}
    \caption{DPTI module: The features are correlated with learnt part-weights to collate the attention vectors and initialize the part-templates.}
    \label{fig:template_module}
\end{figure}

The part templates $\bm{t}_{ i}$ are therefore created from the input data itself, and are able to capture data-specific templates of various human parts that are later used to search for and aggregate discriminative human part features. Our proposed scheme, as illustrated in Fig. \ref{fig:template_module}, reduces the complexity of doing multiple linear projections of keys and queries. It also allows for faster and easier optimization of the part-weights $\bm{q}_{t, i}$.

\subsection{Part Attention Module}

The Part Attention Module is required to localize and aggregate discriminative information about each part of the person identified by the templates $\bm{t}_i$, $i =  1, \dots, N$.
To maximize discriminative power of the model, we use the feature maps from the final layer $\mathbf{F}$ to create the final part vectors. Similar to the template creation, we first pass $\mathbf{F}$ through a $1 \times 1$ convolutional layer to reduce the feature dimensions from $n$ to $d$ and obtain $\tilde{\mathbf{F}}$. Then, following equations \eqref{projection}, \eqref{attention 1}, \eqref{attention 2}, and \eqref{ffn}, we pass $\tilde{\mathbf{F}}$ through a spatial encoder layer and obtain the feature vectors, which are defined as $\mathbf{F}_{a}$. 

The obtained feature vectors are then used as the keys and values, while the part templates are used as the queries of a cross-attention encoder layer. Specifically, the keys, queries, and values are obtained by linear projection of the inputs as given by
\begin{align}
    \bm{q}_{i} &= \mathbf{W}_q \bm{f}_{t, i}, &
    \mathbf{K} &= \mathbf{W}_k \mathbf{F}_{a}, &
    \mathbf{V} &= \mathbf{W}_v \mathbf{F}_{a},
    \label{projection_2}
\end{align}
where $\mathbf{W}_q, \mathbf{W}_k, \mathbf{W}_v \in \mathbb{R}^{d \times d}$ are the linear projection matrices. The attention outputs are obtained following equations \eqref{attention 1} and \eqref{attention 2}, and the final part vectors are obtained after they are passed through a feed-forward network, as in equation \eqref{ffn}. The final output of this is $N$ part vectors $\bm{f}_{p, i}$, $i = 1, \dots, N$. 

The global descriptor vector is obtained by applying global average pooling (GAP) over all outputs of the spatial encoder layer, $\bm{f}_{a, i}, i = 1, \dots, hw$. 

Finally, all $N$ part vectors and the global descriptor vector are concatenated to form the final descriptor vector $\bm{f} \in \mathbb{R}^{(N+1)d \times 1}$. 

\subsection{Adaptive Part Weighting}
\label{sec:adaptive weighting}

Occluded Re-ID and Partial Re-ID both violate the assumption that all localized parts of the human body will be present in the person image. Dynamically initialized part templates will provide some robustness when all parts are not present. To further strengthen robustness in the presence of occlusions, we explore adaptive weighting of the part vectors, $\bm{f}_{p, i}$ before the final concatenation. In particular, we obtain the attention weights that are computed during the template initialization phase in Eq. \eqref{part weights}, $\mathbf{A} = [\bm{\gamma}_{1}, \dots, \bm{\gamma}_{N}] \in \mathbb{R}^{h_l w_l \times N}$. The weight matrix $\mathbf{A}$ represents the correlation between the features obtained from the intermediate layer, and the part-weights. The mean magnitude of each of $\bm{a}_i$ represent the level of activation for the $i$-th part, indicating whether the part is present or not. To leverage this information, we compute the mean of each weight vector $\bm{\gamma}_i$ as $\Bar{\gamma}_i$. We apply a softmax operation over all $N$ mean weights, resulting in a part-presence vector $\bm{\rho} = [\rho_1, \dots, \rho_N]$. Scaling $\bm{\rho}$ by a factor of $N$, we obtain an adaptive weight for each of the parts. Then, before concatenating the part descriptors, we scale each part descriptor by its corresponding weight, as
\begin{equation}
    \tilde{\bm{f}}_{p, i} = N \rho_{i} \bm{f}_{p, i}~.
\end{equation}

For lower values of $\rho_{i}$ to have lower impact on the final distance between descriptors, we use the cosine distance metric after normalization. 

\subsection{Loss Function}

We apply a weighted combination of cross-entropy loss $\lossL_{class}$, triplet loss $\lossL_{tri}$, and a modified diversity loss \cite{li2021diverse} $\lossL_{div}$. Most works that use part descriptors apply the weighted loss separately to each of the part descriptors and the global descriptor. However, we find that concatenating all descriptors into a single descriptor and jointly optimizing using identity classification loss and triplet loss yields superior results. The loss function can be written as
\begin{equation}
  \lossL = \alpha \lossL_{class} + \beta \lossL_{tri} + \gamma \lossL_{div},
  \label{eq:part_loss}
\end{equation}
where $\alpha, \beta$ and $\gamma$ are the weighting factors of each of the losses.

\noindent \textbf{Modified Diversity Loss:} The diversity loss that we propose can be written as:
\begin{equation}
    \lossL_{div} = \frac{1}{N (N-1)} \sum_{i=1}^{N} 
                    \sum_{\substack{j=1\\ j \neq i}}^{N}
                    \frac{\langle \bm{f}_{p,i}, \bm{f}_{p, j} \rangle ^2 }
                    {||\bm{f}_{p,i}||^2_2 ||\bm{f}_{p, j}||^2_2}~. 
\end{equation}
We square each inner product to coerce the part-vectors to be orthogonal, as opposed to its original form presented in \cite{li2021diverse} that allows the vectors to negatively correlated. 
Negative correlation still implies linear dependence, leading to part vectors which may not be independent. 

\section{Experiments}
\renewcommand*{\thefootnote}{\fnsymbol{footnote}}

\subsection{Datasets and Evaluation Metrics}
To demonstrate the effectiveness of our method, we conduct extensive experiments on one occluded dataset: Occluded REID \cite{zhuo2018occluded_Occluded-REID}, two partial Re-ID datasets: Partial-REID \cite{zheng2015partial_Partial-REID} and Partial-iLIDS \cite{zheng2011person_iLIDS}, and one holistic Re-ID dataset: Market-1501 \cite{zheng2015scalable_Market-1501}. The details are as follows.\\
\textbf{Occluded-REID} \cite{zhuo2018occluded_Occluded-REID} is an occluded person dataset captured
by mobile cameras, including 2,000 images belonging to 200 identities. Each identity has five full-body person images and five occluded person images with different viewpoints and different types of severe occlusions.\\
\textbf{Partial-REID} \cite{zheng2015partial_Partial-REID} is a specially designed partial person
Re-ID benchmark that includes 600 images from 60 people, with five full-body images in gallery set and five partial images in query set per person.\\
\textbf{Partial-iLIDS} \cite{he2018deep_Partial-iLIDS} is a partial person Re-ID dataset based
on the iLIDS dataset \cite{zheng2011person_iLIDS} and contains a total of 238 images from 119 people captured by multiple cameras in the airport, and their occluded regions are manually cropped. \\
\textbf{Market-1501} \cite{zheng2015scalable_Market-1501} consists of 1,501 identities captured by
6 cameras. The training set consists of 12,936 images of 751 identities, the query set consists of 3,368 images, and the gallery set consists of 19,732 images.\\
\textbf{Evaluation Metrics:} We adopt standard metrics as in most person Re-ID literature, namely Rank-1 accuracy and mean average precision (mAP), to evaluate the quality of different Re-ID models. Additionally, for our own comparisons in Sec. \ref{sec:ablation}, we report Rank-5 accuracy, and also use the mean Inverse Negative Penalty (mINP) metric introduced in \cite{ye2021deep} to provide a comprehensive overview of the performance of the models.

\subsection{Implementation Details}
We conduct all experiments using PyTorch \cite{NEURIPS2019_9015} with a single 16GB NVIDIA Quadro RTX 5000 GPU. Our architecture is implemented on top of the FastReID toolbox \cite{he2020fastreid}. All pedestrian images are resized to a resolution of $256 \times 128$. We use horizontal flipping, random erasing, padding, and Augmix \cite{hendrycks2019augmix} as data augmentation. All models are trained for 180 epochs on the Market1501 dataset, using Adam optimizer with the base learning rate set to $3.5 \times 10^{-4}$. We adopt linear warm-up for the first 10 epochs, increasing the learning rate from $3.5 \times 10^{-5}$. We decay the learning rate by a factor of 10 at epochs 40, 90, and 150. During training, all feed-forward layers use dropout with probability 0.1, and all attention modules use dropout with probability 0.2. For occluded and partial datasets, random cropping is also utilized to prevent overfitting. None of the models compared with state of the art results include the adaptive weighting module.

For all datasets, the number of part vectors $N$ is set to 8. The attention modules are all implemented as multi-headed attention with 8 heads. The embedding dimension $d$ is set to 256. We use a ResNet50 model pretrained on ImageNet, with the last stride set to 1, and choose the intermediate layer $l$ as 3 to utilize the mid-level features that represents human part-level information. 

\subsection{Comparison with the state-of-the-art}

\begin{table}[ht!]
    \centering
    \begin{tabular}{@{}lcr@{}}
         \toprule
  Method &Rank-1   &   mAP \\
  \toprule
  PCB \cite{sun2018beyond}           & 92.3 & 77.4 \\
  BOT \cite{luo2019bag}           & 94.1 & 85.7 \\
  MGN \cite{wang2018learning}           & 95.7 & 86.9 \\
  VPM \cite{sun2019perceive}           & 93.0 & 80.8 \\
  IANet \cite{hou2019interaction}         & 94.4 & 83.1  \\
  SPReID \cite{kalayeh2018human}        & 92.5 & 81.3 \\
  \footnotemark[2]DSA-reID \cite{zhang2019densely}      & 95.7 & 87.6 \\
  PGFA \cite{miao2019pose}          & 91.2 & 76.8\\
  HOReID \cite{wang2020high}        & 94.2 & 84.9 \\
  RGA-SC \cite{Zhang_2020_CVPR}& 96.1 & 88.4 \\
  $\dagger$ EXAM \cite{qi2021exam}& 95.1 & 85.9 \\
  $\dagger$ PAT \cite{li2021diverse}& 95.4 & 88.0 \\
  $\dagger$ HLGAT \cite{Zhang_2021_CVPR}& 97.5 & 93.4 \\
  \midrule
  \textbf{OURS}         & 94.9 & 86.3 \\
  \bottomrule
    \end{tabular}
    \caption{Comparison with state of the art methods on Market1501 dataset.}
    \label{tab:market-1501}
\end{table}

\noindent \textbf{Results on Holistic Datasets:} The results that we obtain on Market1501 are presented in Table \ref{tab:market-1501}. Our method outperforms many recent attempts, and compares favourably with other state of the art architectures. However, our proposed architecture does not require any handcrafting, and therefore requires minimal intervention by way of network design. It is interesting to note that both HLGAT \cite{Zhang_2021_CVPR}, which far outperforms all other methods, and RGA-SC \cite{Zhang_2020_CVPR} are both complementary to our architecture, and can be implemented in combination. In general, they are valuable for any part-based architecture. However, the method proposed in \cite{Zhang_2021_CVPR} presents significant run-time challenges, as each unique pair of descriptors must be passed through a Graph Attention Network to compute the final embedding. In general, our model is able to achieve very competitive results while making very few assumptions about the nature of the input data.

\begin{table}[ht!]
    \centering
    \begin{tabular}{@{}llr@{}}
         \toprule
         Method &Rank-1   &   mAP \\
  \toprule
  PCB \cite{sun2018beyond}           & 14.3  & 38.9\\
  AMC+SWM \cite{zheng2015partial_Partial-REID}       & 31.2  & 27.3\\
  DSR \cite{he2018deep}           & 72.8  & 62.8\\
  FPR \cite{he2019foreground}           & 78.3  & 68.0\\
  PVPM \cite{gao2020pose}          & 70.4  & 61.2\\
  GASM \cite{he2020guided}          & 74.5  & 65.6\\
  HOReID \cite{wang2020high}        & 80.3  & 70.2\\
  $\dagger$ PAT \cite{li2021diverse}           & 81.6  & 72.1\\
  SSGR \cite{Yan_2021_ICCV}          & 78.5  & 72.9\\
  $\dagger$ OAMN \cite{Chen_2021_ICCV}          & 62.6 & 46.1\\
  \midrule
  \textbf{OURS}         &  71.3   & 67.4\\
  \bottomrule
    \end{tabular}
    \caption{Comparison with state of the art methods on Occluded ReID dataset.}
    \label{tab:occluded_reid}
\end{table}

\noindent \textbf{Results on Occluded Datasets:} 
The results we obtain on Occluded ReID are presented in Table \ref{tab:occluded_reid}. Our method outperforms several recent work while providing competitive perform with the rest. Unlike DSR \cite{he2018deep} and FPR \cite{he2019foreground} that use multiple feature extractors to match the occluded with holistic images, the proposed dynamic part initialization method learns occlusion aware features that provides close results to them.
\footnotetext[2]{indicates that code is not publicly available and results could not be reproduced.}
Our model is capable of providing required local features that are needed to learn the topology of a human unlike HOReID \cite{wang2020high} that uses external human parsing model to do so. PAT \cite{li2021diverse} achieves state of the art results on the Occluded Re-ID dataset. However, the static templates are specialized for a given dataset that raises questions about the model's ability to generalize to other scenarios. Conversely, our templates are created from the source pedestrian image itself, leading to less training bias and greater ability to generalize. Further, unlike \cite{li2021diverse} that requires alteration of model templates to be learnt, our proposed approach provides a unique model that extracts the most prominent attributes for a given input image using mid-level features. 

\begin{table}[ht!]
    \centering
    \small
    \begin{tabular}{cccp{0.05cm}cc}
         \toprule
         \multirow{2}{*}{Method} &
            \multicolumn{2}{c}{Partial-ReID} &&
  \multicolumn{2}{c}{Partial-iLIDS} \\
  \cmidrule{2-3} \cmidrule{5-6}
  &Rank-1   &   Rank-3 && Rank-1 & Rank-3\\
  \toprule
  AMC+SWM \cite{zheng2015partial_Partial-REID}  & 37.3  & 46.0  & & 21.0  &32.8 \\
  DSR \cite{he2018deep}   & 50.7  & 70.0  & & 58.8  & 67.2 \\
  $\dagger$ STNReID \cite{luo2020stnreid}  & 66.7  & 80.3   & & 54.6   &71.3 \\
  VPM \cite{sun2019perceive}  & 67.7  & 81.9  & & 65.5  & 74.8 \\
  PVPM \cite{gao2020pose} & 78.3 & 87.7  & &- & - \\
  HOReID \cite{wang2020high}  &85.3 & 91.0 & & 72.6  &86.4 \\
  $\dagger$ PAT \cite{li2021diverse}  &88.0  &92.3  & &76.5 & 88.2\\
  $\dagger$ PPCL \cite{he2021partial}  &79.0  &87.3  & & - & -\\
  \midrule
  \textbf{OURS}  &  81.3     &     89.0  &&   65.6    & 74.8     \\
  \bottomrule
    \end{tabular}
    \caption{Comparison with state of the art methods on Partial-ReID and Partial-iLIDS datasets.}
    \label{tab:partial}
\end{table}

\noindent \textbf{Results on Partial Datasets:} 
The results we obtain on Partial ReID are presented in Table \ref{tab:partial}. Our proposed architecture outperforms almost every past work except HOReID \cite{wang2020high} and PAT \cite{li2021diverse} on both the Partial-ReID and Partial-iLIDS datasets. On Partial-ReID dataset, it is notable that our method performs competitively well with the state-of-the-art. Unlike \cite{wang2020high}, \cite{li2021diverse}, we do not either rely on parsing models and graph-like topologies to match partial human images or change initialized templates depending on the dataset. This is further reflected on the results we obtained on Partial-iLIDS.  

\subsection{Ablation}
\label{sec:ablation}


\begin{table*}[ht!]
    \centering
    \small
    \begin{tabular}{c p{0.1cm} c c c c p{0.1cm} c c c c p{0.1cm} c c c c}
    \toprule
         Variation &&  \multicolumn{4}{c}{Market-1501}&& \multicolumn{4}{c}{Occluded Re-ID} && \multicolumn{4}{c}{Partial Re-ID} \\
         \cmidrule{3-6} \cmidrule{8-11} \cmidrule{13-16}
         && R-1 & R-5 & mAP & mINP && R-1 & R-5 & mAP & mINP && R-1 & R-5 & mAP & mINP\\
         \toprule
         G &&  93.4 &	97.9 &	82.9 &	53.6            && 59.5 & 75.4 & 52.8 & 38.0 && 79.0 & 90.3 & 72.4 & 58.8\\
         G + P &&  92.3	&  97.5  &	79.6 &	43.4        && 69.9 & 84.6 & 63.5 & 49.4 && 47.3 & 68.7 & 44.9 & 32.9\\
         G + T &&  94.0	& 98.1 &	87.0 &	62.2        && 70.2 & 83.8 & 66.0 & 53.6 && 78.0 & 89.7 & 72.3 & 59.3\\
         G + D &&  95.0 & 98.2 & 86.3 &	60.7            && 71.3 & 86.4 & 67.4 & 55.1 && 81.3 & 91.7 & 76.2 & 64.2\\
         G + D + W &&  94.7 &	98.3  &	85.8 &	59.4    && 71.9 & 85.7 & 66.3 & 53.1 && 81.0 & 90.7 & 76.0 & 63.5\\
         \bottomrule
    \end{tabular}
    \caption{Comparison of different components of the proposed architecture on three datasets}
    \label{tab:ablation_different_components}
\end{table*}

\begin{table}[ht!]
    \centering
    \small
    \begin{tabular}{c c c c c c}
    \toprule
         Backbone & R-1 & R-5 & R-10 & mAP & mINP\\
         \toprule
         Resnet50 &	94.7  &	98.3 &	98.9 &	85.8 &	59.4 \\
         ResNet50 with IBN  &	94.8 &	98.2 &	99.0 &	87.1 &	63.1 \\
         OSNet      & 94.4 &   98.3 & 98.2 & 86.7   & 62.3 \\
         ResNet18 &	93.1 &	97.9 &	98.9 &	81.7 &	50.2 \\
         \bottomrule
    \end{tabular}
    \caption{Results of the proposed architecture using various backbones on Market 1501.}
    \label{tab:backbone}
\vspace{-4mm}
\end{table}

\begin{figure}
    \centering
    \includegraphics[width=\linewidth]{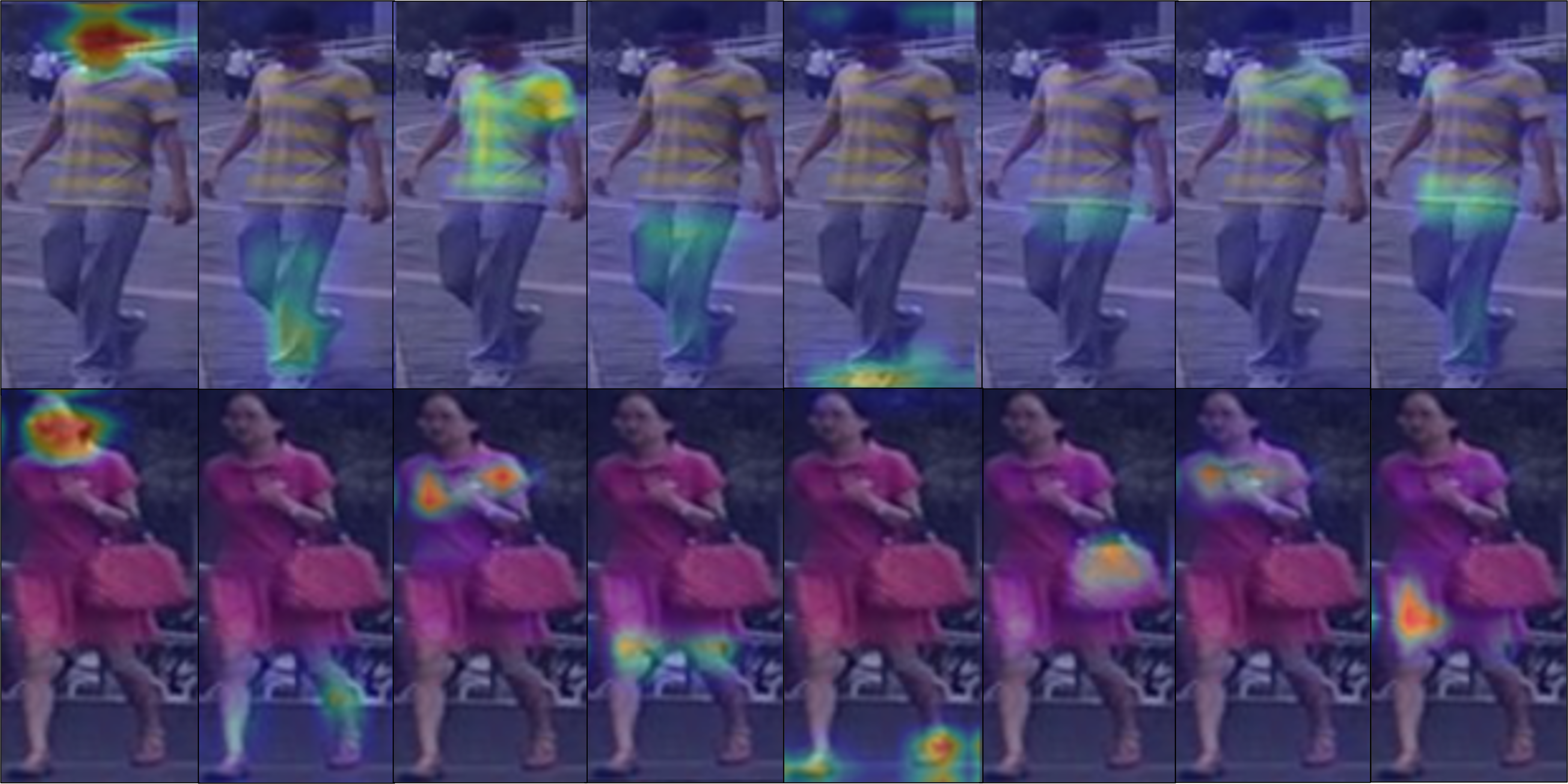}
    \caption{Attention maps computed by the template initialization layer. Images are taken from the Market1501 dataset.}
    \label{fig:attention output}
\end{figure}

\noindent \textbf{Effect of DPTI: } We perform ablation studies to determine the impact and effectiveness of each of the components of the proposed architecture. We compare variations of the proposed DPTI module against a baseline which uses only the global features obtained through a self-attention layer (G). To prove the effectiveness of our dynamic initialization, we introduce 2 variations. The first is to remove the template initialization module, and partition the attention outputs $\bm{f}_{a_l, i}$ into a grid of $N$ sections, and average each section into a single feature vector which is used as the part template (P). This demonstrates the importance of dynamically localizing parts, instead of making assumptions about their location in the image. The second is to use static templates instead of dynamically initialized templates (T). This demonstrates the importance of data-specific templates. Our proposed model uses dynamically initialized templates (D). Finally, we also present the results when the final feature vectors are weighted using the adaptive weighting scheme (W) introduced in Sec. \ref{sec:adaptive weighting}. The results are presented in Table \ref{tab:ablation_different_components}. 
As demonstrated by the results presented in the table, using averaged features from the intermediate layer (P) alone impacts performance negatively, especially for partial Re-ID. This is because the model assumes that a given part only occurs in a given section of the pedestrian image. Using static templates (T) increases performance. However, without dynamic initialization, the templates cannot adjust to differences in input data, and will only reflect the characteristics of the training dataset. As such, the static template initialization method is able to perform at a similar level for holistic Re-ID, but performance drops off for Occluded and Partial Re-ID. It is interesting to note that the adaptive weighting module (W) does not produce any significant impact on performance, and in fact, slightly degrades performance for Occluded Re-ID. We include it only as an exploration of adaptive weighting.

\noindent \textbf{Effect of backbone architecture:} We also evaluate our architecture using different backbones on Market 1501 to demonstrate that our architecture is able to leverage the feature extraction properties of various different models. We use a small model, ResNet18, and models of different structures, such as ResNet50 with IBN \cite{pan2018two}, and OSNet \cite{zhou2019omni}. The results are presented in Table \ref{tab:backbone}. Impressively, the ResNet18 model with our proposed architecture is able to achieve a mINP score of 50.2 \%, despite its small size. Additionally, our architecture is able to adapt well to different backbones, and utilize their feature extraction capabilities. Both the ResNet50 with IBN and the OSNetx1.0 achieve similar results to that of the ResNet50, with the ResNet50 with IBN unsurprisingly surpassing the results obtained by the baseline ResNet50, due to its superior generalization abilities. 

\noindent \textbf{Qualitative Results:} We plot the attention maps from the template initialization layer to conduct a qualitative investigation into what the part-templates are focusing on. The results are plotted in Fig. \ref{fig:attention output}. As shown by the plot, each template is paying attention to a particular part of the human body, including the head, torso, shoes, and even various accessories being carried by the pedestrian. These are generally useful parts to focus on for Re-ID of a person. It is intuitive that the model would consider these features when determining what parts to localize. 
\subsection{Limitations and Negative Social Impacts}

Results presented by most Person Re-ID architectures are limited by the data available for training. High mAP and Rank-1 accuracy scores can only be achieved under the conditions present in the dataset. One significant limitation of our work is the lack of training datasets. With the termination of both DukeMTMC-Re-ID \cite{ristani2016performance, zhuo2018occluded_Occluded-REID} due to privacy concerns \cite{Exposing.ai}, and the MSMT17 \cite{wei2018person} from the original website for unknown reasons, we felt it was best to not use those datasets. However, we are confident that the principles introduced are transferable to most applications of Person Re-ID, and many other applications including vehicle Re-ID and multi-camera tracking. Additionally, many existing Person Re-ID datasets are too small to successfully train attention modules. 

There are many assumptions associated with closed-world Person Re-ID, such as well-aligned bounding boxes, sufficient illumination, and a closed gallery set. These do not hold true in real-world applications. However, these are limitations of closed-world Person Re-ID, and we do not consider them as limitations of our contribution.

A serious ethical concern for collecting large Person Re-ID datasets is the privacy of the human participants, as exemplified by the DukeMTMC-Re-ID termination. 
As observed by \cite{Exposing.ai}, DukeMTMC-Re-ID continues to be used both for research and commercial applications. Another issue which is largely ignored by researchers is the biases that are introduced by these datasets into the models, which are subsequently used for commercial applications. Deep investigation into these biases should be encouraged before research commences on any dataset.

Overall, the Re-ID domain has serious societal implications. Increasing public surveillance worldwide is driven by the techniques being developed herein. While there are definite security benefits to intelligent surveillance, it may also be put to ill-use, or become compromised due to unauthorized access by malicious parties. Additionally, inaccuracy in Re-ID models could lead to mistaken identity, which may have severe consequences. Thus, human verification in the loop is essential, as is moving towards more secure, accurate models for responsible Person Re-ID.
\section{Conclusion}
In this work, we introduce a novel dynamic part template initialization module, specially for partial and occluded person Re-ID, to extract part-based spatial templates by utilizing mid-level features extracted at middle layers of the backbone. We obtain background noise-reduced attention vectors by passing the mid-level feature maps through a self-attention layer followed by our cross-attention-inspired projection to initialize part templates. To extract the discriminative part-descriptors with only identity labels, we propose a part attention module and modified diversity loss which are effective in discriminating local feature embeddings from high-level feature maps. We also analyzed the use of an adaptive part-weighting mechanism that weighs the part-descriptors based on the mean activation of the part template-attention weights. The qualitative and quantitative results show that the dynamically initialed templates learns to localize human parts and provides competitive results without using additional feature extractors or dataset dependent part-initialization. 

Our model can be further extended by utilizing the part-descriptors to compute higher order topological structures for better representation or utilizing backbones with stronger representation power. Further, implementation of a distance metric that is able to utilize the adaptive weighting module may improve occluded and partial Re-ID performance. The dynamic local feature initialization can be generalized to many other vision tasks, such as object detection during autonomous driving, by modeling and propagating important local features at earlier layers of a backbone that leads to understand challenging scenes better.

{\small
\bibliographystyle{ieee_fullname}
\bibliography{egbib}

\begin{thebibliography}{10}\itemsep=-1pt

\bibitem{Chen_2019_ICCV}
Binghui Chen, Weihong Deng, and Jiani Hu.
\newblock Mixed high-order attention network for person re-identification.
\newblock In {\em Proceedings of the IEEE/CVF International Conference on
  Computer Vision}, October 2019.

\bibitem{Chen_2021_ICCV}
Peixian Chen, Wenfeng Liu, Pingyang Dai, Jianzhuang Liu, Qixiang Ye, Mingliang
  Xu, Qi{\textquoteright}an Chen, and Rongrong Ji.
\newblock Occlude them all: Occlusion-aware attention network for occluded
  person re-id.
\newblock In {\em Proceedings of the IEEE/CVF International Conference on
  Computer Vision}, pages 11833--11842, October 2021.

\bibitem{chen2017beyond}
Weihua Chen, Xiaotang Chen, Jianguo Zhang, and Kaiqi Huang.
\newblock Beyond triplet loss: a deep quadruplet network for person
  re-identification.
\newblock In {\em Proceedings of the IEEE conference on computer vision and
  pattern recognition}, pages 403--412, 2017.

\bibitem{fan2018scpnet}
Xing Fan, Hao Luo, Xuan Zhang, Lingxiao He, Chi Zhang, and Wei Jiang.
\newblock Scpnet: Spatial-channel parallelism network for joint holistic and
  partial person re-identification.
\newblock In {\em Asian conference on computer vision}, pages 19--34. Springer,
  2018.

\bibitem{gao2020pose}
Shang Gao, Jingya Wang, Huchuan Lu, and Zimo Liu.
\newblock Pose-guided visible part matching for occluded person reid.
\newblock In {\em Proceedings of the IEEE/CVF Conference on Computer Vision and
  Pattern Recognition}, pages 11744--11752, 2020.

\bibitem{gong_xiang_2011}
Shaogang Gong and Tao Xiang.
\newblock {\em Visual analysis of behaviour: From pixels to semantics}.
\newblock Springer, 2011.

\bibitem{Exposing.ai}
Jules. Harvey, Adam.~LaPlace.
\newblock Exposing.ai, 2021.

\bibitem{he2016deep}
Kaiming He, Xiangyu Zhang, Shaoqing Ren, and Jian Sun.
\newblock Deep residual learning for image recognition.
\newblock In {\em Proceedings of the IEEE conference on computer vision and
  pattern recognition}, pages 770--778, 2016.

\bibitem{he2018deep}
Lingxiao He, Jian Liang, Haiqing Li, and Zhenan Sun.
\newblock Deep spatial feature reconstruction for partial person
  re-identification: Alignment-free approach.
\newblock In {\em Proceedings of the IEEE Conference on Computer Vision and
  Pattern Recognition}, pages 7073--7082, 2018.

\bibitem{he2018deep_Partial-iLIDS}
Lingxiao He, Jian Liang, Haiqing Li, and Zhenan Sun.
\newblock Deep spatial feature reconstruction for partial person
  re-identification: Alignment-free approach.
\newblock In {\em Proceedings of the IEEE Conference on Computer Vision and
  Pattern Recognition}, pages 7073--7082, 2018.

\bibitem{he2020fastreid}
Lingxiao He, Xingyu Liao, Wu Liu, Xinchen Liu, Peng Cheng, and Tao Mei.
\newblock Fastreid: A pytorch toolbox for general instance re-identification.
\newblock {\em arXiv preprint arXiv:2006.02631}, 2020.

\bibitem{he2020guided}
Lingxiao He and Wu Liu.
\newblock Guided saliency feature learning for person re-identification in
  crowded scenes.
\newblock In {\em European Conference on Computer Vision}, pages 357--373.
  Springer, 2020.

\bibitem{he2018recognizing}
Lingxiao He, Zhenan Sun, Yuhao Zhu, and Yunbo Wang.
\newblock Recognizing partial biometric patterns.
\newblock {\em arXiv preprint arXiv:1810.07399}, 2018.

\bibitem{he2019foreground}
Lingxiao He, Yinggang Wang, Wu Liu, He Zhao, Zhenan Sun, and Jiashi Feng.
\newblock Foreground-aware pyramid reconstruction for alignment-free occluded
  person re-identification.
\newblock In {\em Proceedings of the IEEE/CVF International Conference on
  Computer Vision}, pages 8450--8459, 2019.

\bibitem{he2021partial}
Tianyu He, Xu Shen, Jianqiang Huang, Zhibo Chen, and Xian-Sheng Hua.
\newblock Partial person re-identification with part-part correspondence
  learning.
\newblock In {\em Proceedings of the IEEE/CVF Conference on Computer Vision and
  Pattern Recognition}, pages 9105--9115, 2021.

\bibitem{hendrycks2019augmix}
Dan Hendrycks, Norman Mu, Ekin~D Cubuk, Barret Zoph, Justin Gilmer, and Balaji
  Lakshminarayanan.
\newblock Augmix: A simple data processing method to improve robustness and
  uncertainty.
\newblock {\em arXiv preprint arXiv:1912.02781}, 2019.

\bibitem{hou2019interaction}
Ruibing Hou, Bingpeng Ma, Hong Chang, Xinqian Gu, Shiguang Shan, and Xilin
  Chen.
\newblock Interaction-and-aggregation network for person re-identification.
\newblock In {\em Proceedings of the IEEE/CVF Conference on Computer Vision and
  Pattern Recognition}, pages 9317--9326, 2019.

\bibitem{kalayeh2018human}
Mahdi~M Kalayeh, Emrah Basaran, Muhittin G{\"o}kmen, Mustafa~E Kamasak, and
  Mubarak Shah.
\newblock Human semantic parsing for person re-identification.
\newblock In {\em Proceedings of the IEEE conference on computer vision and
  pattern recognition}, pages 1062--1071, 2018.

\bibitem{koestinger2012large}
Martin Koestinger, Martin Hirzer, Paul Wohlhart, Peter~M Roth, and Horst
  Bischof.
\newblock Large scale metric learning from equivalence constraints.
\newblock In {\em 2012 IEEE conference on computer vision and pattern
  recognition}, pages 2288--2295. IEEE, 2012.

\bibitem{kolesnikov2021image}
Alexander Kolesnikov, Alexey Dosovitskiy, Dirk Weissenborn, Georg Heigold,
  Jakob Uszkoreit, Lucas Beyer, Matthias Minderer, Mostafa Dehghani, Neil
  Houlsby, Sylvain Gelly, et~al.
\newblock An image is worth 16x16 words: Transformers for image recognition at
  scale.
\newblock In {\em International Conference on Learning Representations}, 2021.

\bibitem{li2018harmonious}
Wei Li, Xiatian Zhu, and Shaogang Gong.
\newblock Harmonious attention network for person re-identification.
\newblock In {\em Proceedings of the IEEE conference on computer vision and
  pattern recognition}, pages 2285--2294, 2018.

\bibitem{li2021diverse}
Yulin Li, Jianfeng He, Tianzhu Zhang, Xiang Liu, Yongdong Zhang, and Feng Wu.
\newblock Diverse part discovery: Occluded person re-identification with
  part-aware transformer.
\newblock In {\em Proceedings of the IEEE/CVF Conference on Computer Vision and
  Pattern Recognition}, pages 2898--2907, 2021.

\bibitem{li2018unified}
Yaoyu Li, Tianzhu Zhang, Lingyu Duan, and Changsheng Xu.
\newblock A unified generative adversarial framework for image generation and
  person re-identification.
\newblock In {\em Proceedings of the 26th ACM international conference on
  Multimedia}, pages 163--172, 2018.

\bibitem{liao2015person}
Shengcai Liao, Yang Hu, Xiangyu Zhu, and Stan~Z Li.
\newblock Person re-identification by local maximal occurrence representation
  and metric learning.
\newblock In {\em Proceedings of the IEEE conference on computer vision and
  pattern recognition}, pages 2197--2206, 2015.

\bibitem{liu2018pose}
Jinxian Liu, Bingbing Ni, Yichao Yan, Peng Zhou, Shuo Cheng, and Jianguo Hu.
\newblock Pose transferrable person re-identification.
\newblock In {\em Proceedings of the IEEE Conference on Computer Vision and
  Pattern Recognition}, pages 4099--4108, 2018.

\bibitem{luo2019bag}
Hao Luo, Youzhi Gu, Xingyu Liao, Shenqi Lai, and Wei Jiang.
\newblock Bag of tricks and a strong baseline for deep person
  re-identification.
\newblock In {\em Proceedings of the IEEE/CVF Conference on Computer Vision and
  Pattern Recognition Workshops}, pages 0--0, 2019.

\bibitem{luo2020stnreid}
Hao Luo, Wei Jiang, Xing Fan, and Chi Zhang.
\newblock Stnreid: Deep convolutional networks with pairwise spatial
  transformer networks for partial person re-identification.
\newblock {\em IEEE Transactions on Multimedia}, 22(11):2905--2913, 2020.

\bibitem{matsukawa2016hierarchical}
Tetsu Matsukawa, Takahiro Okabe, Einoshin Suzuki, and Yoichi Sato.
\newblock Hierarchical gaussian descriptor for person re-identification.
\newblock In {\em Proceedings of the IEEE conference on computer vision and
  pattern recognition}, pages 1363--1372, 2016.

\bibitem{miao2019pose_Occludded-Duke}
Jiaxu Miao, Yu Wu, Ping Liu, Yuhang Ding, and Yi Yang.
\newblock Pose-guided feature alignment for occluded person re-identification.
\newblock In {\em Proceedings of the IEEE/CVF International Conference on
  Computer Vision}, pages 542--551, 2019.

\bibitem{miao2019pose}
Jiaxu Miao, Yu Wu, Ping Liu, Yuhang Ding, and Yi Yang.
\newblock Pose-guided feature alignment for occluded person re-identification.
\newblock In {\em Proceedings of the IEEE/CVF International Conference on
  Computer Vision}, pages 542--551, 2019.

\bibitem{pan2018two}
Xingang Pan, Ping Luo, Jianping Shi, and Xiaoou Tang.
\newblock Two at once: Enhancing learning and generalization capacities via
  ibn-net.
\newblock In {\em Proceedings of the European Conference on Computer Vision
  (ECCV)}, pages 464--479, 2018.

\bibitem{NEURIPS2019_9015}
Adam Paszke, Sam Gross, Francisco Massa, Adam Lerer, James Bradbury, Gregory
  Chanan, Trevor Killeen, Zeming Lin, Natalia Gimelshein, Luca Antiga, Alban
  Desmaison, Andreas Kopf, Edward Yang, Zachary DeVito, Martin Raison, Alykhan
  Tejani, Sasank Chilamkurthy, Benoit Steiner, Lu Fang, Junjie Bai, and Soumith
  Chintala.
\newblock Pytorch: An imperative style, high-performance deep learning library.
\newblock In H. Wallach, H. Larochelle, A. Beygelzimer, F. d\textquotesingle
  Alch\'{e}-Buc, E. Fox, and R. Garnett, editors, {\em Advances in Neural
  Information Processing Systems 32}, pages 8024--8035. Curran Associates,
  Inc., 2019.

\bibitem{qi2021exam}
Guanqiu Qi, Gang Hu, Xiaofei Wang, Neal Mazur, Zhiqin Zhu, and Matthew Haner.
\newblock Exam: A framework of learning extreme and moderate embeddings for
  person re-id.
\newblock {\em Journal of Imaging}, 7(1):6, 2021.

\bibitem{ristani2016performance}
Ergys Ristani, Francesco Solera, Roger Zou, Rita Cucchiara, and Carlo Tomasi.
\newblock Performance measures and a data set for multi-target, multi-camera
  tracking.
\newblock In {\em European conference on computer vision}, pages 17--35.
  Springer, 2016.

\bibitem{sarfraz2018pose}
M~Saquib Sarfraz, Arne Schumann, Andreas Eberle, and Rainer Stiefelhagen.
\newblock A pose-sensitive embedding for person re-identification with expanded
  cross neighborhood re-ranking.
\newblock In {\em Proceedings of the IEEE Conference on Computer Vision and
  Pattern Recognition}, pages 420--429, 2018.

\bibitem{song2018mask}
Chunfeng Song, Yan Huang, Wanli Ouyang, and Liang Wang.
\newblock Mask-guided contrastive attention model for person re-identification.
\newblock In {\em Proceedings of the IEEE conference on computer vision and
  pattern recognition}, pages 1179--1188, 2018.

\bibitem{su2017pose}
Chi Su, Jianing Li, Shiliang Zhang, Junliang Xing, Wen Gao, and Qi Tian.
\newblock Pose-driven deep convolutional model for person re-identification.
\newblock In {\em Proceedings of the IEEE international conference on computer
  vision}, pages 3960--3969, 2017.

\bibitem{sun2019perceive}
Yifan Sun, Qin Xu, Yali Li, Chi Zhang, Yikang Li, Shengjin Wang, and Jian Sun.
\newblock Perceive where to focus: Learning visibility-aware part-level
  features for partial person re-identification.
\newblock In {\em Proceedings of the IEEE/CVF Conference on Computer Vision and
  Pattern Recognition}, pages 393--402, 2019.

\bibitem{sun2018beyond}
Yifan Sun, Liang Zheng, Yi Yang, Qi Tian, and Shengjin Wang.
\newblock Beyond part models: Person retrieval with refined part pooling (and a
  strong convolutional baseline).
\newblock In {\em Proceedings of the European conference on computer vision
  (ECCV)}, pages 480--496, 2018.

\bibitem{vaswani2017attention}
Ashish Vaswani, Noam Shazeer, Niki Parmar, Jakob Uszkoreit, Llion Jones,
  Aidan~N Gomez, {\L}ukasz Kaiser, and Illia Polosukhin.
\newblock Attention is all you need.
\newblock In {\em Advances in neural information processing systems}, pages
  5998--6008, 2017.

\bibitem{wang2020high}
Guan'an Wang, Shuo Yang, Huanyu Liu, Zhicheng Wang, Yang Yang, Shuliang Wang,
  Gang Yu, Erjin Zhou, and Jian Sun.
\newblock High-order information matters: Learning relation and topology for
  occluded person re-identification.
\newblock In {\em Proceedings of the IEEE/CVF Conference on Computer Vision and
  Pattern Recognition}, pages 6449--6458, 2020.

\bibitem{wang2018learning}
Guanshuo Wang, Yufeng Yuan, Xiong Chen, Jiwei Li, and Xi Zhou.
\newblock Learning discriminative features with multiple granularities for
  person re-identification.
\newblock In {\em Proceedings of the 26th ACM international conference on
  Multimedia}, pages 274--282, 2018.

\bibitem{wei2018person}
Longhui Wei, Shiliang Zhang, Wen Gao, and Qi Tian.
\newblock Person transfer gan to bridge domain gap for person
  re-identification.
\newblock In {\em Proceedings of the IEEE conference on computer vision and
  pattern recognition}, pages 79--88, 2018.

\bibitem{wolfe1992parallel}
Jeremy~M Wolfe.
\newblock The parallel guidance of visual attention.
\newblock {\em Current Directions in Psychological Science}, 1(4):124--128,
  1992.

\bibitem{xiong2014person}
Fei Xiong, Mengran Gou, Octavia Camps, and Mario Sznaier.
\newblock Person re-identification using kernel-based metric learning methods.
\newblock In {\em European conference on computer vision}, pages 1--16.
  Springer, 2014.

\bibitem{Yan_2021_ICCV}
Cheng Yan, Guansong Pang, Jile Jiao, Xiao Bai, Xuetao Feng, and Chunhua Shen.
\newblock Occluded person re-identification with single-scale global
  representations.
\newblock In {\em Proceedings of the IEEE/CVF International Conference on
  Computer Vision}, pages 11875--11884, October 2021.

\bibitem{yang2014salient}
Yang Yang, Jimei Yang, Junjie Yan, Shengcai Liao, Dong Yi, and Stan~Z Li.
\newblock Salient color names for person re-identification.
\newblock In {\em European conference on computer vision}, pages 536--551.
  Springer, 2014.

\bibitem{ye2021deep}
Mang Ye, Jianbing Shen, Gaojie Lin, Tao Xiang, Ling Shao, and Steven~CH Hoi.
\newblock Deep learning for person re-identification: A survey and outlook.
\newblock {\em IEEE Transactions on Pattern Analysis and Machine Intelligence},
  2021.

\bibitem{zhang2018learning}
Tianzhu Zhang, Changsheng Xu, and Ming-Hsuan Yang.
\newblock Learning multi-task correlation particle filters for visual tracking.
\newblock {\em IEEE transactions on pattern analysis and machine intelligence},
  41(2):365--378, 2018.

\bibitem{zhang2019densely}
Zhizheng Zhang, Cuiling Lan, Wenjun Zeng, and Zhibo Chen.
\newblock Densely semantically aligned person re-identification.
\newblock In {\em Proceedings of the IEEE/CVF Conference on Computer Vision and
  Pattern Recognition}, pages 667--676, 2019.

\bibitem{Zhang_2020_CVPR}
Zhizheng Zhang, Cuiling Lan, Wenjun Zeng, Xin Jin, and Zhibo Chen.
\newblock Relation-aware global attention for person re-identification.
\newblock In {\em IEEE/CVF Conference on Computer Vision and Pattern
  Recognition}, June 2020.

\bibitem{Zhang_2021_CVPR}
Zhong Zhang, Haijia Zhang, and Shuang Liu.
\newblock Person re-identification using heterogeneous local graph attention
  networks.
\newblock In {\em Proceedings of the IEEE/CVF Conference on Computer Vision and
  Pattern Recognition}, pages 12136--12145, June 2021.

\bibitem{zhao2017deeply}
Liming Zhao, Xi Li, Yueting Zhuang, and Jingdong Wang.
\newblock Deeply-learned part-aligned representations for person
  re-identification.
\newblock In {\em Proceedings of the IEEE international conference on computer
  vision}, pages 3219--3228, 2017.

\bibitem{zheng2015scalable}
Liang Zheng, Liyue Shen, Lu Tian, Shengjin Wang, Jingdong Wang, and Qi Tian.
\newblock Scalable person re-identification: A benchmark.
\newblock In {\em Proceedings of the IEEE international conference on computer
  vision}, pages 1116--1124, 2015.

\bibitem{zheng2015scalable_Market-1501}
Liang Zheng, Liyue Shen, Lu Tian, Shengjin Wang, Jingdong Wang, and Qi Tian.
\newblock Scalable person re-identification: A benchmark.
\newblock In {\em Proceedings of the IEEE international conference on computer
  vision}, pages 1116--1124, 2015.

\bibitem{zheng2016person}
Liang Zheng, Yi Yang, and Alexander~G Hauptmann.
\newblock Person re-identification: Past, present and future.
\newblock {\em arXiv preprint arXiv:1610.02984}, 2016.

\bibitem{zheng2011person_iLIDS}
Wei-Shi Zheng, Shaogang Gong, and Tao Xiang.
\newblock Person re-identification by probabilistic relative distance
  comparison.
\newblock In {\em CVPR 2011}, pages 649--656. IEEE, 2011.

\bibitem{zheng2012reidentification}
Wei-Shi Zheng, Shaogang Gong, and Tao Xiang.
\newblock Reidentification by relative distance comparison.
\newblock {\em IEEE transactions on pattern analysis and machine intelligence},
  35(3):653--668, 2012.

\bibitem{zheng2015partial_Partial-REID}
Wei-Shi Zheng, Xiang Li, Tao Xiang, Shengcai Liao, Jianhuang Lai, and Shaogang
  Gong.
\newblock Partial person re-identification.
\newblock In {\em Proceedings of the IEEE International Conference on Computer
  Vision}, pages 4678--4686, 2015.

\bibitem{zhou2019omni}
Kaiyang Zhou, Yongxin Yang, Andrea Cavallaro, and Tao Xiang.
\newblock Omni-scale feature learning for person re-identification.
\newblock In {\em Proceedings of the IEEE/CVF International Conference on
  Computer Vision}, pages 3702--3712, 2019.

\bibitem{zhuo2018occluded_Occluded-REID}
Jiaxuan Zhuo, Zeyu Chen, Jianhuang Lai, and Guangcong Wang.
\newblock Occluded person re-identification.
\newblock In {\em 2018 IEEE International Conference on Multimedia and Expo
  (ICME)}, pages 1--6. IEEE, 2018.

\bibitem{zhuo2019novel}
Jiaxuan Zhuo, Jianhuang Lai, and Peijia Chen.
\newblock A novel teacher-student learning framework for occluded person
  re-identification.
\newblock {\em arXiv preprint arXiv:1907.03253}, 2019.

\end{thebibliography}
}

\end{document}